\pdfoutput=1
\documentclass[10pt, a4paper]{article}
\usepackage{lrec}
\usepackage{multibib}
\newcites{languageresource}{Language Resources}
\usepackage{graphicx}
\usepackage{tabularx}
\usepackage{soul}
\usepackage{epstopdf}

\usepackage{hyperref}
\usepackage{xstring}
\usepackage{xspace}
\usepackage{color}
\usepackage{enumitem}
\usepackage[usenames, dvipsnames]{xcolor}
\usepackage{booktabs}
\usepackage{amsfonts}
\usepackage{multirow}
\usepackage{comment}
\usepackage[utf8]{inputenc}
\usepackage[normalem]{ulem}
\useunder{\uline}{\ul}{}

\title{EmotionLines: An Emotion Corpus of Multi-Party Conversations}

\name{Sheng-Yeh Chen$^{1*}$, Chao-Chun Hsu$^{1*}$, Chuan-Chun Kuo$^1$,\\{\bf \large Ting-Hao (Kenneth) Huang$^2$, Lun-Wei Ku$^1$}}

\address{Academia Sinica, Taiwan \\
         Carnegie Mellon University, USA \\
         $^1$\{alan012102, joe32140, william0617, lwku\}@iis.sinica.edu.tw\\
         $^2$tinghaoh@cs.cmu.edu\\}

\abstract{
Feeling emotion is a critical characteristic to distinguish people from machines. Among all the multi-modal resources for emotion detection, textual datasets are those containing the least additional information in addition to semantics, and hence are adopted widely for testing the developed systems. However, most of the textual emotional datasets consist of emotion labels of only individual words, sentences or documents, which makes it challenging to discuss the contextual flow of emotions. In this paper, we introduce EmotionLines, the first dataset with emotions labeling on all utterances in each dialogue only based on their textual content.
Dialogues in EmotionLines are collected from Friends TV scripts and private Facebook messenger dialogues. Then one of seven emotions, six Ekman's basic emotions plus the neutral emotion, is labeled on each utterance by 5 Amazon MTurkers. A total of 29,245 utterances from 2,000 dialogues are labeled in EmotionLines. We also provide several strong baselines for emotion detection models on EmotionLines in this paper. \\ \newline \Keywords{emotion detection, emotional dialogue dataset}}

\begin{document}
\maketitleabstract
\section{Introduction}
\noindent
There are two major kinds of dialogue systems: a task-oriented dialogue system and the social (chit-chat) dialogue system. The former focuses on designing a personal assistant which can accomplish certain tasks, and for the latter it is important to capture the conversation flow which emphasizes more on the feelings of the speaker. Many researchers try to build a ``smart'' dialogue system by enhancing dialogue breadth (coverage), dialogue depth (complexity) or both. Those who want to increase dialogue breadth try to transfer dialogue acts across domains~\cite{chen2016zero} to establish multi-domain or even open domain dialogue system, and those who want to deepen dialogue complexity pay their attention to transform a knowledge-based systems to common sense or even empathetic systems that can recognize emotion features, generate emotion-aware responses~\cite{fung2016towards}, or learn how to plan the dialogues while users interact via high-level descriptions~\cite{sun2016appdialogue}. No matter what kind of dialogue system we want to build, a useful and large dialogue dataset is indispensable.{\let\thefootnote\relax\footnote{{$^{*}$These authors contributed equally to this work.}}}

When building a task-oriented dialogue system, dialogue corpora with dialogue act information is accessible and hence are commonly utilized. However when building a chit-chat conversational bot, though the importance of emotion detection has been noticed, pure conversation content such as movie, TV scripts or chat logs without emotion labels are more available: no emotion labels on utterances can be used for learning. Moreover, when we turn to other datasets with annotated emotion information such as data crawled from Twitter~\cite{mohammad2017wassa}, the labeled units (posts or sentences) are independent. As a result, models built with these datasets lack the ability to consider contextual information essential in dialogue systems, not to mention the ability to capture the emotion flow. We illustrate this issue with examples shown in Table~\ref{tab:intro1}.

\begin{table}[ht!]
\centering
%\footnotesize
\begin{tabular}{lc}
\toprule
\multicolumn{1}{c}{\textbf{Post}} & \textbf{Label} \\ \midrule
\begin{tabular}[c]{@{}l@{}}Just got back from seeing @GaryDelaney in \\Burslem. AMAZING!! Face still hurts from \\laughing so much \#hilarious\end{tabular} & Joy \\ \midrule
Feeling worthless as always \#depression & Sadness \\ \midrule
\begin{tabular}[c]{@{}l@{}}I get so nervous even thinking about talking\\ to *** *** I wanna die\end{tabular} & Fear \\ \midrule
\begin{tabular}[c]{@{}l@{}}Wont use using @mothercareuk\\ @Mothercarehelp again!! These guys cant\\ get nothing right!! \#fuming\end{tabular} & Anger \\ \bottomrule
\end{tabular}
\caption{Emotion labeled posts without contextual information (selected from WASSA-2017 Shared Task on Emotion Intensity)}
\label{tab:intro1}
\end{table}

Modeling emotion on one single utterance without contextual information may encounter another issue that the same utterance can express different emotions depending on its context. Table~\ref{tab:intro2} shows some examples of saying ``Okay!'' with different emotions.

\begin{table}[ht!]
\centering
%\footnotesize
\begin{tabular}{cl}
\toprule
\textbf{Chandler} & \begin{tabular}[c]{@{}l@{}}Matthew Perry talking about signs in Las\\ Vegas. (Neutral)\end{tabular} \\
\textbf{Chandler} & \begin{tabular}[c]{@{}l@{}}I guess it must've been some movie I saw.\\ (Neutral)\end{tabular} \\
\textbf{Chandler} & What do you say? (Neutral) \\
\textbf{Monica} & \textbf{\textit{Okay! (Joy)}} \\
\textbf{Chandler} & Okay! Come on! Let's go! All right! (Joy) \\ \midrule
\textbf{Rachel} & \begin{tabular}[c]{@{}l@{}}Oh okay, I'll fix that to. What's her e-mail\\ address? (Neutral)\end{tabular} \\
\textbf{Ross} & Rachel! (Anger) \\
\textbf{Rachel} & \begin{tabular}[c]{@{}l@{}}All right, I promise. I'll fix this. I swear. \\I'll-I'll- I'll-I'll talk to her. (Non-neutral)\end{tabular} \\
\textbf{Ross} & \textbf{\textit{Okay! (Anger)}} \\
\textbf{Rachel} & Okay. (Neutral) \\ \bottomrule
\end{tabular}
\caption{``Okay!'' of different emotions from Friends TV scripts.}
\label{tab:intro2}
\vspace{1.0pc}
\end{table}
%\shengyeh{examples of dialogue data and emotion data}

% Sentiment analysis has become a key to set up a successful dialogue system as it enables systems to capture human feelings and react like humans. Opinion analysis and emotion recognition are two major tasks in sentiment analysis. Specially on text materials, the former derives benefits from easily obtained gold labels generally from social media, while the latter suffers from the lack of obvious resources and relies heavily on manual annotation, which intensify the difficulty of emotion recognition on text.
The IEMOCAP database~\cite{busso2008iemocap}, to the best of our knowledge, is the only dataset that provides emotion labels for each utterance. However, IEMOCAP was created by 
actors performing emotions, and hence carries the risk of overacting. Moreover, the annotators label the emotions by watching the videos instead of reading the transcripts which means the annotators may make the decision only depend on the facial expression or the prosodic features without realizing the meaning of the words.

To tackle these problems, we create EmotionLines: an emotion dialogue dataset with emotion labels on each utterance. The collected textual dialogues are from not only scripts of TV shows but also real, private, human-to-human chat logs. We establish several strong baselines for the emotion detection task on dialogues, and motivate an automatic metric to benchmark progress. Modeling sequential emotions in dialogues, as provided in EmotionLines, has the potential to move dialog systems from generating understandable messages to more human-like responses. To the best of our knowledge, this is the first textual emotion dialogue dataset with the emotion label on each utterance in dialogues. 

\section{Related Work}
\noindent
Sentiment analysis, which can help users and companies capture people's opinion, is getting a growing attention by both research community and business word because of the research challenge and the potential value to make profits. Since the social media and the instant message platforms become the important part of our daily life, we can easily obtain a large amount of these user-generated content to get better understanding of emotion. Thus improve the satisfaction for web services and call centers~\cite{devillers2006real}.

In 1974, Ekman conducted extensive studies on emotion recognition research over 6 basic emotions: anger, disgust,  fear, happiness, sadness, and surprise. His study shows it is possible to detect emotions given enough features ~\cite{ekman1987universals}. Later studies on text-based emotion recognition are mainly divided into three categories: keyword-based, learning-based, and hybrid recommendation approaches~\cite{kao2009textbased}. Recently, emotion recognition researches on text focus on the learning-based methods. Kim proposed CNN(Convolutional neural network) text classification, which is widely used for extracting sentence information~\cite{kim2014convolutional}. However, single sentence emotion recognition is lack of contextual emotion flow within a dialogue. Therefore, contextual LSTM(Long short-term memory) architecture is proposed to measure the inter-dependency of utterances in the dialogue~\cite{soujanyaacl17}. In this paper, we report the performance of the CNN model and the contextual LSTM architecture on the proposed EmotionLines dataset as baselines.
  
% sentiment analysis
% single sentence
% conversation 
% Ekman emotion
%
%
%
%
%
\section{Corpus}

\subsection{Data Source}
\noindent
To bring conversations closer to real-word dialogues, we selected sources from both TV shows scripts and Human-to-human chat logs. First, we crawled the scripts of seasons 1 to 9 of \textit{Friends} TV shows\footnote[1]{Scripts of seasons 1-9 of ``Friends'': \url{http://www.livesinabox.com/friends/scripts.shtml}}. Second, we requested private dialogues from Wang~\shortcite{wang2016sensing}, which are conversations between friends on Facebook Messenger collected by an app called \textit{EmotionPush} \footnote[2]{Participants consented to
make their private conversations available for research purposes.}.

\subsubsection{Friends TV Scripts}
\noindent
The crawled scripts are separated as episodes, and we viewed each scene in every episode as a dialogue. Then, the collected dialogues were categorized according to their dialogue length, i.e. the number of utterances in a dialogue, into four classes of which bucket length ranges are [5, 9], [10, 14], [15, 19], and [20, 24]. Finally, we randomly sampled 250 dialogues from each class to construct a dataset containing 1,000 dialogues.

\subsubsection{EmotionPush Chat Logs}
\noindent
For the private dialogues from \textit{EmotionPush}, we assumed that a dialogue would not sustain more than 30 minutes, and messages separated in time by less than 300 seconds were put in the same dialogue. At last, the dialogues are categorized and sampled using the same procedure as that for the Friends TV scripts, and we obtained 1,000 dialogues from EmotionPush chat logs.

\subsection{Human-level Labeling}
\begin{figure}[t]
\begin{center}
%\fbox{\parbox{6cm}{
%This is a figure with a caption.}}
\includegraphics[width=\columnwidth]{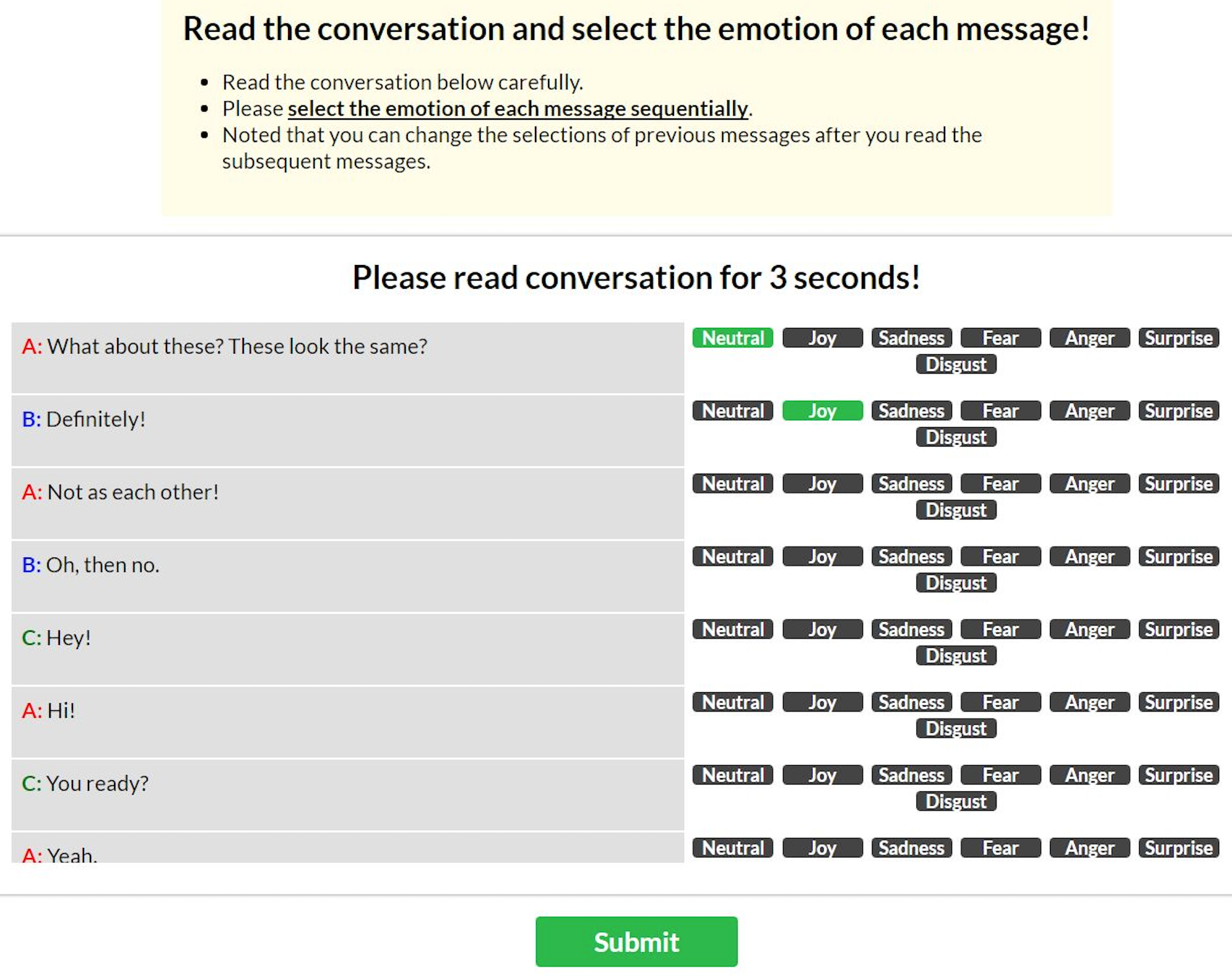} 
\caption{Worker interface on Amazon Mechanical Turk}
\label{fig.1}
\end{center}
\end{figure}

\noindent
We placed our dialogues on the Amazon Mechanical Turk, and each dialogue is regarded as an annotation task where each utterance is labeled with one of Ekman's six basic emotions~\shortcite{ekman1987universals} anger, disgust,fear,  happiness,  sadness, surprise, and the additional emotion neutral. The total of seven labels are Neutral, Joy, Sadness, Fear, Anger, Surprise, and Disgust respectively. For every MTurk HIT, we designed a web interface like Figure~\ref{fig.1}, and asked crowd workers to mark each utterance in a dialogue considering the context in the whole dialogue. Workers should think for at least 3 seconds before selecting an answer. For HITs with different dialogue length, we assign distinctive payments according to the bucket length ranges mentioned above, where the award is 0.1, 0.15, 0.2, and 0.25 dollars per HIT respectively.

Each HIT was accomplished by 5 workers, and for each utterance in a HIT, the emotion with the highest number of votes was set as the gold label of the utterance. Those utterances with more than two different emotions voted were put into the non-neutral category.

\subsection{De-identification}
%\paragraph{Named Entity Recognition and Replacement}
The EmotionPush chat logs are from private conversations. Therefore, the logs may contain personal information such as names of real people, locations, organizations, and email addresses. In order to protect the privacy of EmotionPush users, we performed a two-step masking process. First, Stanford Named Entity Recognizer~\cite{finkel2005incorporating} was adopted to detect named entities mentioned by each utterance, which were later replaced with their entity types.

After this step, however, we still found named entities like lowercase or foreign names and emails. Therefore, we manually checked and cleaned utterances in the first step again to prevent the accidental reveal of personal data.

%\paragraph{Rewriting Messages}
Since the conversations collected in the EmotionPush chat logs involve not only our participants, we want to carefully protect their identity. Therefore, we hired a native speaker whose occupation is an editor to rewrite all the messages of participants' friends. The rewriting process follows the guideline from Bruckman~\cite{bruckman2006teaching}. In addition, we asked the rewriter to check all named entities and mask them by its categories. For example, ``my mother'' remains the same, but ``Sam'' is replaced by ``person $n$'' indicating the $n^{th}$ person being de-identified; ``the park near my house'' remains the same but ``Taipei 101'' is replaced by ``location $m$'' indicating the $m^{th}$ location being de-identified.

\subsection{Data Format}
\noindent
Both Friends TV scripts and EmotionPush chat logs contain 1,000 dialogues. The lengths of dialogues vary. Each utterance of the dialogue has the same format, which involves information of the speaker, the content, and the emotion labels of the utterance. We show an example in Table~\ref{tab:data-format}.

\begin{table}[t]
\centering
%\footnotesize
\begin{tabular*}{\columnwidth}{cl}
\toprule
\textbf{speaker} & Rachel \\
\textbf{utterance} & Hi Joey! What are you doing here? \\
\textbf{emotion} & joy \\ \midrule
\textbf{speaker} & Joey \\
\textbf{utterance} & \begin{tabular}[c]{@{}l@{}}Uhh, well I've got an audition down \\the street and I spilled sauce \\all over the front of my shirt. \\You got an extra one?\end{tabular} \\
\textbf{emotion} & neutral \\ \midrule
\textbf{speaker} & Rachel \\
\textbf{utterance} & Yeah, sure. Umm... here. \\
\textbf{emotion} & neutral \\ \bottomrule
\end{tabular*}
\caption{Data format of EmotionLines}
\label{tab:data-format}
\end{table}

\section{Analysis}

\subsection{Data Information}
\noindent
The analysis of data from two sources is shown in Table~\ref{tab:data-information}. We calculated the numbers of utterances of each type of emotion, and found that except \textit{neutral} type, \textit{joy} and \textit{surprise} appear more frequently in the dataset. Besides, EmotionPush chat logs have more skewed label distribution than Friends TV scripts.
Interestingly, the average length of real private utterances is much shorter than the length of those of TV show scripts (10.67 vs. 6.84).

\begin{table*}[ht!]
\centering
%\footnotesize
\begin{tabular}{c|cc|cccccccc|c}
\toprule
\multirow{2}{*}{} & \multirow{2}{*}{\textbf{\begin{tabular}[c]{@{}c@{}}\# of\\ Utterances\end{tabular}}} & \multirow{2}{*}{\textbf{\begin{tabular}[c]{@{}c@{}}Utterance\\ Length\end{tabular}}} & \multicolumn{8}{c|}{\textbf{Emotion Label Distribution (\%)}} & \multirow{2}{*}{\textbf{\begin{tabular}[c]{@{}c@{}}kappa\\ (\%)\end{tabular}}} \\
 &  &  & Neu & Joy & Sad & Fea & Ang & Sur & Dis & Non &  \\ \midrule
\textbf{Friends} & 14,503  & 10.67 & 45.03 & 11.79 & 3.43 & 1.70 & 5.23 & 11.43 & 2.28 & 19.11 & 33.83 \\
\textbf{EmotionPush} & 14,742  & 6.84 & 66.85 & 14.25 & 3.49 & 0.28 & 0.95 & 3.85 & 0.72 & 9.62 & 33.64 \\ \bottomrule
\end{tabular}
\caption{Detail information of Friends TV scripts and EmotionPush chat logs}
\label{tab:data-information}
\end{table*}

% \usepackage{booktabs}
% \usepackage{multirow}
% \usepackage{graphicx}
% \usepackage[normalem]{ulem}
\begin{table*}[]
\centering
%\footnotesize
%\resizebox{\textwidth}{!}
{%
\begin{tabular}{@{}llccccccccc@{}}
\toprule

\multicolumn{1}{c}{} & \multicolumn{1}{c}{{\ul }} & \textbf{WA} & \textbf{UWA} & Neu & Joy & Sad & Fea & Ang & Sur & Dis\\ \midrule
\multicolumn{1}{l|}{\multirow{2}{*}{CNN}} & \textbf{Friends} & \textbf{59.2} & \multicolumn{1}{c|}{\textbf{45.2}} & 64.3 & 60.2 & 41.2 & 21.9 & 46.6 & 61.5 & 20.6 \\  
\multicolumn{1}{l|}{} & \textbf{EmotionPush$^{*}$} & \textbf{71.5} & \multicolumn{1}{c|}{\textbf{41.7}} & 80.8 & 46.9 & 43.7 & 0.0 & 27.0 & 53.8 & 40.0 \\ \midrule
\multicolumn{1}{l|}{\multirow{2}{*}{CNN-BiLSTM}} & \textbf{Friends} & \textbf{63.9} & \multicolumn{1}{c|}{\textbf{43.1}} & 74.7 & 61.8 & 45.9 & 12.5 & 46.6 & 51.0 & 8.8\\ 
\multicolumn{1}{l|}{} & \textbf{EmotionPush$^{*}$} & \textbf{77.4} & \multicolumn{1}{c|}{\textbf{39.4}} & 87.0 & 60.3 & 28.7 & 0.0 & 32.4 & 40.9 & 26.7  \\ \bottomrule
\end{tabular}%
}
\caption{Weighted and unweighted accuracy on Friends and EmotionPush}
\label{tab:performance}
\end{table*}

%\begin{table}[ht!]
\begin{table*}[ht!]
\centering
%\footnotesize
\begin{tabular}{c|ccc|ccc}
\hline
\multirow{2}{*}{} & \multicolumn{3}{c|}{\textbf{Friends}} & \multicolumn{3}{c}{\textbf{EmotionPush}} \\
 & train & dev & test & train & dev & test \\ \cline{2-7} 
\textbf{\# of D} & 720 & 80 & 200 & 720 & 80 & 200 \\
\textbf{\# of U} & 10,561 & 1,178 & 2,764 & 10,733 & 1,202 & 2,807 \\
\textbf{D-len} & 14.67 & 14.73 & 13.82 & 14.91 & 15.03 & 14.04 \\
\textbf{U-len} & 10.20 & 10.08 & 10.44 & 6.73 & 6.96 & 7.24 \\ \hline
\end{tabular}
\caption{Information of train/dev/test set of Friends and EmotionPush dataset (D and U represent dialogue and utterance correspondingly, and the dialogue/utterance lengths were averaged.)}
\label{tab:train_test_split}
\end{table*}

We adopted Fleiss' kappa to measure the agreement among annotators of the labeling task of the dataset. The kappa scores are above 0.33 for labels of both the Friends scripts and EmotionPush, which indicates a solid basis for a subjective labeling task.

\subsection{Train-/Dev-/Test-Set Split}
\noindent
We not only constructed an emotion dialogue corpus, but also split the dataset from two sources into training, development, and testing set separately. In order to preserve completeness of any dialogue, we divided the corpus by the dialogues, not the utterances. Table~\ref{tab:train_test_split} shows the information of each set.

\section{Experiments}

\subsection{Modeling a Single Utterance}
\noindent
Given a utterance of $M$ words, the one-hot encoding for utterance words is denoted by $U$ = \{$w_1$,
$w_2$, $w_3$,..., $w_M$\}. We first embed the words to the word embedding , which is publicly available 300-dimensional GloVe pre-trained on Common Crawl data~\cite{pennington2014glove}. Thus each utterance in $u_i$ is represented by
a feature matrix $F \in \mathbb{R}^{M \times 300}$. Then, a 1-D convolution with $k$ filters of $r$ window sizes from 1 to $r$, followed by a 1-D max-pooling is applied on $F$. The concatenation of max-pooling outputs of different window sizes is denoted as $f$ with dimension $k \times r$. $k$ is set to 64 and $r$ is set to 5 in the experiment.

%\subsubsection{LSTM}
 %We apply the LSTM, long short-term memory, to capture the utterance-level information. LSTM is capable of modeling long-range dependencies, which other traditional RNNs fail to do given the problem of vanishing gradient~\cite{Hochreiter1997LSTM}. We use the last hidden state of the LSTM as utterance-level representation, which is denoted as $h$. The hidden state dimension is set to 512.
 
\subsection{Modeling on the Whole Dialogue}
\noindent
In a paragraph, the sentiment of each utterance is dependent on the context. Thus, within a dialogue, there is a high probability of inter-dependency with respect to their sentimental clues. When we classify an utterance, other utterances may provide import contextual information. To measure this information flow, we apply the contextual LSTM architecture. The inputs of contextual LSTM for each dialogue with length $L$ are denoted as $X$ = \{$x_1$, ..., $x_L$\}.
\begin{equation}
x_i=\tanh(W_x\cdot f_i + b_x)
%\vspace{.1cm}
\end{equation}
The output of LSTM cell $h_i$ is then fed to the dense layer followed by a softmax layer. Then we compute loss by cross-entropy as follows:
\begin{equation}
loss = -\frac{1}{{\sum_{c\in C}N_j }}{\sum_{c\in C}\sum_{i=1}^{N_c}\sum_{l\in C}(y_{l}^{i})log(\hat{y}_{l}^{i})}
\end{equation}
where $C$ is the emotion class set for evaluation, $N_c$ denotes the number of
utterances in class $c$, $y_l^i$ is the original output, and $\hat{y}_l^i$ is the
predicted output for the $i$-th utterance in emotion class $l$.

\subsection{Performance on EmotionLines}
\noindent
We conduct experiments on EmotionLines with the CNN model and the CNN-Bidirectional LSTM(CNN-BiLSTM) model. Results are shown in Table~\ref{tab:performance}. The performance of the CNN and CNN-BiLSTM model is evaluated by both the weighted accuracy (WA) and the unweighted accuracy (UWA) shown as follows.

\begin{equation}
\textnormal{WA} = {\sum_{l \in C}{s}_{l}{a}_{l}}
%\vspace{.1cm}
\end{equation}
\begin{equation}
\textnormal{UWA} = \frac{1}{|C|}{\sum_{l \in C}{a}_{l}}
%\vspace{.1cm}
\end{equation}
where $a_l$ denotes the accuracy of emotion class $l$ and $s_l$ denotes the percentage of utterances in emotion class $l$.

The improvements of weighted accuracy from 59.2\% to 63.9\% on the Friends dataset and from 71.5\% to 77.4\% on the EmotionPush dataset show that using the contextual information (CNN-BiLSTM) can help recognize emotions. Note that the reported performance is from the experiments conducted on the raw data, which are not de-identified yet. Updated results will be provided later in the dataset download webpage\footnote[3]{http://academiasinicanlplab.github.io/\#download}.

%\begin{figure}
%\begin{center}
%\fbox{\parbox{6cm}{
%This is a figure with a caption.}}
%\scalebox{.3}{\includegraphics{figure/friends_5emo.jpg}}
%\includegraphics[width=\columnwidth]{figure/friends_5emo.jpg} 
%\caption{Friends confusion matrix by CNN-BiLSTM}
%\label{fig.2}
%\end{center}
%\end{figure}

%\begin{figure}
%\begin{center}
%\fbox{\parbox{6cm}{
%This is a figure with a caption.}}
%\scalebox{.3}{\includegraphics{figure/emo_push_5emo.jpg} }
%\caption{EmotionPush confusion matrix by CNN-BiLSTM}
%\label{fig.3}
%\end{center}
%\end{figure}

\section{Conclusion and Future Work}
\noindent
We have constructed EmotionLines, the emotion dialogue dataset containing the text content for each utterance annotated with one of seven emotion-categorical labels. The kappa value shows the good quality of these generated labels. In addition, several experiments were performed to provide baselines and to show contextual information is beneficial for the dialogue emotion recognition. The provided strong baselines are weighted accuracy 63.9\% and 77.4\% for Friends and EmotionPush, respectively.

Due to the imbalanced nature of emotion label distribution, one of our future work is to collect specific types of the label to enrich the minor emotion categories, e.g., trying horror movies scripts to get more \textit{fear} utterances and tragedies for \textit{sadness} utterances. EmotionLines is now available at \url{http://academiasinicanlplab.github.io/#download}.

\section{Acknowledgement}
\noindent
This research is partially supported by Ministry of Science and Technology, Taiwan, under Grant no. MOST 106-2218-E-002-043-.

\section{References}

\label{main:ref}
\bibliographystyle{lrec}
\bibliography{lrec}

\begin{thebibliography}{}

\bibitem[\protect\citename{Bruckman}2006]{bruckman2006teaching}
Bruckman, A.
\newblock (2006).
\newblock Teaching students to study online communities ethically.
\newblock {\em Journal of Information Ethics}, page~82.

\bibitem[\protect\citename{Busso \bgroup et al.\egroup }2008]{busso2008iemocap}
Busso, C., Bulut, M., Lee, C.-C., Kazemzadeh, A., Mower, E., Kim, S., Chang,
  J.~N., Lee, S., and Narayanan, S.~S.
\newblock (2008).
\newblock Iemocap: Interactive emotional dyadic motion capture database.
\newblock {\em Language resources and evaluation}, 42(4):335.

\bibitem[\protect\citename{Chen \bgroup et al.\egroup }2016]{chen2016zero}
Chen, Y.-N., Hakkani-T{\"u}r, D., and He, X.
\newblock (2016).
\newblock Zero-shot learning of intent embeddings for expansion by
  convolutional deep structured semantic models.
\newblock In {\em Acoustics, Speech and Signal Processing (ICASSP), 2016 IEEE
  International Conference on}, pages 6045--6049. IEEE.

\bibitem[\protect\citename{Devillers and Vidrascu}2006]{devillers2006real}
Devillers, L. and Vidrascu, L.
\newblock (2006).
\newblock Real-life emotions detection with lexical and paralinguistic cues on
  human-human call center dialogs.
\newblock In {\em Ninth International Conference on Spoken Language
  Processing}.

\bibitem[\protect\citename{Ekman \bgroup et al.\egroup
  }1987]{ekman1987universals}
Ekman, P., Friesen, W.~V., O'sullivan, M., Chan, A., Diacoyanni-Tarlatzis, I.,
  Heider, K., Krause, R., LeCompte, W.~A., Pitcairn, T., Ricci-Bitti, P.~E.,
  et~al.
\newblock (1987).
\newblock Universals and cultural differences in the judgments of facial
  expressions of emotion.
\newblock {\em Journal of personality and social psychology}, 53(4):712.

\bibitem[\protect\citename{Finkel \bgroup et al.\egroup
  }2005]{finkel2005incorporating}
Finkel, J.~R., Grenager, T., and Manning, C.
\newblock (2005).
\newblock Incorporating non-local information into information extraction
  systems by gibbs sampling.
\newblock In {\em Proceedings of the 43rd annual meeting on association for
  computational linguistics}, pages 363--370. Association for Computational
  Linguistics.

\bibitem[\protect\citename{Fung \bgroup et al.\egroup }2016]{fung2016towards}
Fung, P., Bertero, D., Wan, Y., Dey, A., Chan, R. H.~Y., Siddique, F.~B., Yang,
  Y., Wu, C.-S., and Lin, R.
\newblock (2016).
\newblock Towards empathetic human-robot interactions.
\newblock {\em arXiv preprint arXiv:1605.04072}.

\bibitem[\protect\citename{Kao \bgroup et al.\egroup }2009]{kao2009textbased}
Kao, E. C.~C., Liu, C.~C., Yang, T.~H., Hsieh, C.~T., and Soo, V.~W.
\newblock (2009).
\newblock Towards text-based emotion detection a survey and possible
  improvements.
\newblock In {\em 2009 International Conference on Information Management and
  Engineering}, pages 70--74, April.

\bibitem[\protect\citename{Kim}2014]{kim2014convolutional}
Kim, Y.
\newblock (2014).
\newblock Convolutional neural networks for sentence classification.
\newblock {\em arXiv preprint arXiv:1408.5882}.

\bibitem[\protect\citename{Mohammad and Bravo-Marquez}2017]{mohammad2017wassa}
Mohammad, S.~M. and Bravo-Marquez, F.
\newblock (2017).
\newblock Wassa-2017 shared task on emotion intensity.
\newblock {\em arXiv preprint arXiv:1708.03700}.

\bibitem[\protect\citename{Pennington \bgroup et al.\egroup
  }2014]{pennington2014glove}
Pennington, J., Socher, R., and Manning, C.~D.
\newblock (2014).
\newblock Glove: Global vectors for word representation.
\newblock In {\em Empirical Methods in Natural Language Processing (EMNLP)},
  pages 1532--1543.

\bibitem[\protect\citename{Poria \bgroup et al.\egroup }2017]{soujanyaacl17}
Poria, S., Cambria, E., Hazarika, D., Mazumder, N., Zadeh, A., and Morency,
  L.-P.
\newblock (2017).
\newblock Context-dependent sentiment analysis in user-generated videos.
\newblock In {\em Association for Computational Linguistics}.

\bibitem[\protect\citename{Sun \bgroup et al.\egroup }2016]{sun2016appdialogue}
Sun, M., Chen, Y.-N., Hua, Z., Tamres-Rudnicky, Y., Dash, A., and Rudnicky,
  A.~I.
\newblock (2016).
\newblock Appdialogue: Multi-app dialogues for intelligent assistants.
\newblock In {\em LREC}.

\bibitem[\protect\citename{Wang \bgroup et al.\egroup }2016]{wang2016sensing}
Wang, S.-M., Li, C.-H., Lo, Y.-C., Huang, T.-H.~K., and Ku, L.-W.
\newblock (2016).
\newblock Sensing emotions in text messages: An application and deployment
  study of emotionpush.
\newblock {\em arXiv preprint arXiv:1610.04758}.

\end{thebibliography}

\end{document}